\documentclass[10pt,twocolumn,letterpaper]{article}

\usepackage{cvpr}              

\usepackage[dvipsnames]{xcolor}

\usepackage{times}
\usepackage{graphicx}
\usepackage{amsmath}
\usepackage{amssymb}
\usepackage{comment}
\usepackage{algorithm}
\usepackage{algpseudocode}

\usepackage{color}
\usepackage{url}
\usepackage{enumitem}
\usepackage{rotating}
\usepackage{bbm}
\usepackage{cases}

\usepackage{multirow}
\usepackage{mathtools}
\usepackage{xcolor,colortbl}
\usepackage{makecell}
\usepackage{booktabs}

\usepackage{listings}
\usepackage{xcolor}  
\newcommand{\calS}{\ensuremath{\mathcal{S}}}
\newcommand{\calC}{\ensuremath{\mathcal{C}}}
\newcommand{\calM}{\ensuremath{\mathcal{M}}}
\newcommand{\calD}{\ensuremath{\mathcal{D}}}

\newcommand\blfootnote[1]{%
  \begingroup
  \renewcommand\thefootnote{}\footnote{#1}%
  \addtocounter{footnote}{-1}%
  \endgroup
}

\definecolor{cvprblue}{rgb}{0.21,0.49,0.74}
\usepackage[pagebackref,breaklinks,colorlinks,citecolor=cvprblue]{hyperref}

\def\paperID{3089} 
\def\confName{CVPR}
\def\confYear{2025}

\title{Tripartite Weight-Space Ensemble for Few-Shot Class-Incremental Learning}

\author{Juntae Lee\hspace{1em} Munawar Hayat\hspace{1em} Sungrack Yun \\
{Qualcomm AI Research$^\dag$} \\ 
{\tt {\small\{juntlee,hayat,sungrack\}@qti.qualcomm.com}}}

\begin{document}
\maketitle

\blfootnote{\hspace{-1.8em}$^\dag$Qualcomm AI Research is an initiative of Qualcomm Technologies, Inc. and/or its subsidiaries.}

\begin{abstract}
Few-shot class incremental learning (FSCIL) enables the continual learning of new concepts with only a few training examples. In FSCIL, the model undergoes substantial updates, making it prone to forgetting previous concepts and overfitting to the limited new examples. Most recent trend is typically to disentangle the learning of the representation from the classification head of the model. A well-generalized feature extractor on the base classes (many examples and many classes) is learned, and then fixed during incremental learning. Arguing that the fixed feature extractor restricts the model's adaptability to new classes, we introduce a novel FSCIL method to effectively address catastrophic forgetting and overfitting issues. Our method enables to seamlessly update the entire model with a few examples. We mainly propose a tripartite weight-space ensemble (Tri-WE). Tri-WE interpolates the base, immediately previous, and current models in weight-space, especially for the classification heads of the models. Then, it collaboratively maintains knowledge from the base and previous models. In addition, we recognize the challenges of distilling generalized representations from the previous model from scarce data. Hence, we suggest a regularization loss term using amplified data knowledge distillation.
Simply intermixing the few-shot data, we can produce richer data enabling the distillation of critical knowledge from the previous model. Consequently, we attain state-of-the-art results on the miniImageNet, CUB200, and CIFAR100 datasets.
\end{abstract}

\section{Introduction}
\label{sec:intro}

The ability of a model to continually adapt with the availability of new data is highly desirable for many real-world applications. Few-shot class-incremental learning (FSCIL) has recently gained significant research attention as evidenced by several studies~\cite{zhang2021few,peng2022few,kalla2022s3c,song2023learning, zhao2023few}, because it offers a promising learning paradigm. It enables the model to flexibly expand its classification capabilities to new concepts using only a few training examples. FSCIL starts by training a model on numerous base classes with abundant labeled data. In each subsequent incremental session, the model encounters only a limited number of labeled data for new classes, along with a single prototype for each of the previously learned classes. Hence, FSCIL faces multiple challenges, including the retention of previously learned knowledge and the prevention of overfitting to the limited examples of new data.

\begin{figure}[t]
    \centering
    \includegraphics[width=8.5cm]{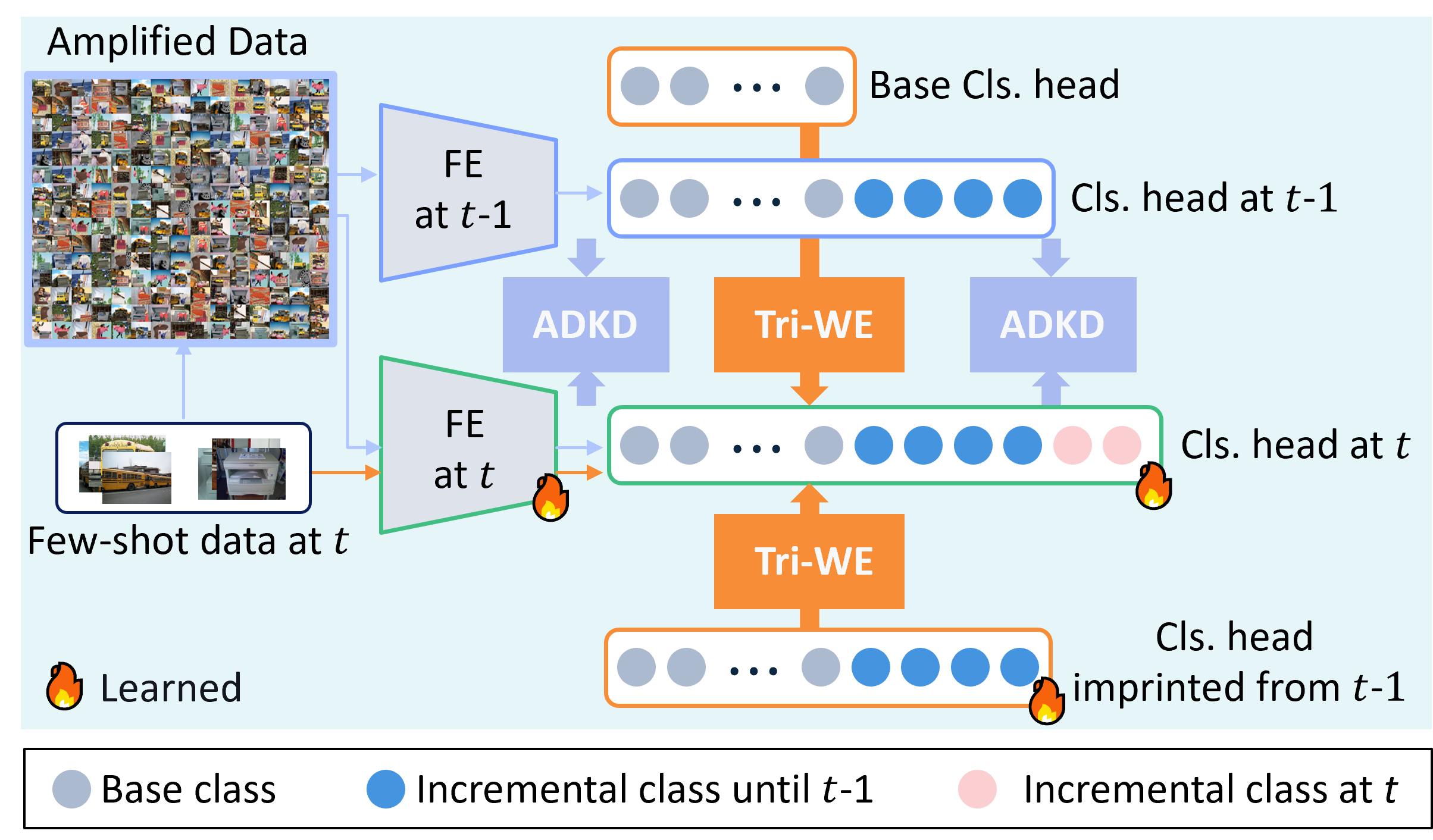}
    \caption{\textbf{Overview of the proposed method} (FE: feature extractor, Cls. head: classification head). Through Tri-WE, at $t$th incremental session, Cls. head is composed by interpolating the weights of the Cls. heads covering the classes until base, $(t-1)$th, $t$th sessions, respectively. For ADKD regularization loss, the input few-shot data are amplified to reliably mine generalized knowledge of the model of the previous $(t-1)$th session. After incremental learning, deployed components are outlined in green.}
    \label{fig:intro}
\end{figure}

\begin{figure}[t]
    \centering
    \includegraphics[width=8.0cm]{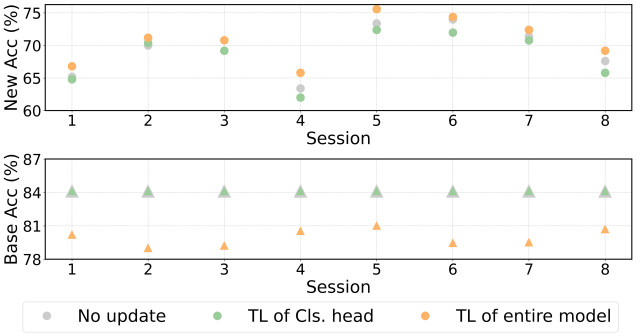}
    \caption{\textbf{Few-shot (10-way 5-shot miniImageNet) transfer learning (TL) starting from the base session model on each session.} Classification accuracy scores on incremented 10 classes (top) \& 40 base classes (bottom) are shown for each session.}
    \label{fig:motiv}
\end{figure}

To address these challenges, 
the most recent works are based on separating learning the representation from the classification. They rely on constructing a robust feature embedding space using plentiful base class data. In subsequent incremental learning where only a few labeled data for new classes are available, the embedding space remains fixed, and the focus is shifted to learning the classifier for the new classes. This strategy is intended to avoid accumulated model collapse or overfitting due to the limited data. For example, \cite{zhang2021few,peng2022few} refrain from altering any base model parameters and use mean features for each class to define their classifiers. In contrast, \cite{kalla2022s3c,song2023learning} employ ensembling multiple classifiers to improve generalization. \cite{zhao2023few} introduces an extra module to assess the reliability of outputs from the base and incrementally learned models for a weighted sum of the two outputs. Moreover, \cite{yang2023neural,ahmed2024orco} fixes the feature embedding space and also pre-defines classifier templates, requiring prior knowledge of the total number of incremental classes to be encountered at the outset of the base class learning. However, as indicated in Fig.~\ref{fig:motiv}, updating the entire model can lead to a drop in base class performance but may enhance the model's adaptability to new classes. Hence, if we address the catastrophic forgetting and overfitting more effectively, we can harness the benefits of updating the entire model.

To this end, we propose a novel framework for FSCIL as depicted in Fig.~\ref{fig:intro}. Mainly, inspired by Wise-FT \cite{wiseft}, we develop \textit{tripartite weight-space ensemble (Tri-WE)} to continually update the weights of the model. We interpolate the classification heads of models from the base, intermediately previous, and current sessions within the weight space, thereby achieving a balance between adaptability to new classes and retention of knowledge of previous ones. This approach allows for a continual update of knowledge without an abrupt shift in the decision boundary. Moreover, to address catastrophic forgetting, knowledge distillation (KD) has been a common regularization technique in class incremental learning. However, its application in a few-shot scenario is prone to overfitting to the few available examples, and then recent cutting-edge methods have less used KD for FSCIL. To counter this, we also introduce \textit{amplified data KD (ADKD) regularization loss} where we intermix few-shot training datasets, enriching the data source for distilling the knowledge from the previous model. It simply ensures that the model prevents catastrophic forgetting by transferring the generalized knowledge for previous classes. We achieve state-of-the-art (SOTA) results on three benchmarks: miniImageNet, CUB200, and CIFAR100.

Our contributions are summarized as three-fold:
\begin{itemize}
    \item We introduce the Tri-WE, wherein the classification heads from base, preceding, and current models are synergistically interpolated within the weight space, promoting a knowledge continuum. 
    \vspace{0.1cm}
    \item Tackling distillation of biased knowledge resulting from a scarcity of data in KD, we develop ADKD sub-loss term that enhances the generality of the distilled knowledge.
    \vspace{0.1cm}

    \item Leveraging these proposed components, we attain SOTA results across on miniImageNet, CUB200, and CIFAR100. Comprehensive experiments demonstrate the significance and efficacy of our contributions.
\end{itemize}

\section{Related Works}
\label{sec:rel}

\noindent\textbf{Class-incremental learning (CIL)} aims to seamlessly integrate new classes without access to previously seen data, preventing the catastrophic forgetting problem. To ensure knowledge retention, rehearsal of selective previous samples~\cite{chaudhry2018efficient,rolnick2019experience} has been developed. Also, in~\cite{wang2022learning,yan2021dynamically}, the network is dynamically expanded to accommodate new learnings. LwF~\cite{li2017learning} and iCaRL~\cite{rebuffi2017icarl} used KD and nearest class mean (NCM) classifiers, respectively, to mitigate the catastrophic forgetting. UCIR~\cite{hou2019learning} introduced a novel feature distillation loss, and PODNet~\cite{douillard2020podnet} extended distillation beyond the final embedding output, incorporating the pooled outputs from the model’s intermediate layers. In~\cite{wu2019large,zhao2020maintaining,belouadah2019il2m}, the class imbalance problem is tackled. However, these methods depend on ample training data, an aspect not feasible in FSCIL, necessitating new strategies that adapt swiftly with minimal data while preserving existing knowledge.

\noindent\textbf{FSCIL} fuses CIL and few-shot learning. It gained attention after TOPIC~\cite{tao2020few} initiated the concept, leading to numerous subsequent methods. Recent works align with two strategies. One approach completely fixes the learnable weights of the model after the base session's many-shot many-class training, and then incrementally extends the classifier by adding mean features of new classes. CEC~\cite{zhang2021few} enhances the base model by simulating the FSCIL scenarios sampled from the base data again, while ALICE~\cite{peng2022few} employs diverse techniques to boost the generalization capability of the base model. Alternatively, some methods~\cite{kalla2022s3c,yang2023neural,song2023learning,zhao2023few} update a part of the feature extractor or the added class-mean-initialized classifiers in each session. SoftNet~\cite{kang2022soft} finds the weights less impactful for the catastrophic forgetting at the base session, and focuses on updating them in the incremental learning. In WaRP~\cite{kim2022warping}, the important weights are decided in every session, and they keep frozen in the following sessions. GKEAL~\cite{zhuang2023gkeal} updates the weight of the classification head using least square solver in the view of analytic learning.
S3C~\cite{kalla2022s3c} employed the stochastic classifier as the extreme ensemble of classifiers. NS-FSCIL~\cite{yang2023neural} and OrCo~\cite{ahmed2024orco} assume the base session knows the amount of total incremental classes and incremental session starts with mapping each class to a classifier predefined in the base session. NS-FSCIL optimizes the last module of the feature encoder and OrCo does the last module and the selected pre-defined classifier. In SAVC~\cite{song2023learning}, a class is recognized by a set of classifiers, each tailored to a specific input augmentation, resulting in a swift expansion of deployed model size during incremental learning. In BiDistill~\cite{zhao2023few}, every prediction is made by the ensemble of the base and last models.  





\noindent\textbf{Network ensemble} has long been an important technique in deep learning in order to reduce overfitting and improve generalization ability~\cite{hansen1990neural,dietterich2000ensemble}. It involves the training of multiple neural networks and combining them for the final prediction. Various ensemble methods have been developed to get different but collaborative models such as random-weight initialization~\cite{krizhevsky2012imagenet}, different training data~\cite{szegedy2015going}, and capturing multiple snapshots in training schedules~\cite{huang2017iclr}. While these methods have consistently demonstrated improved performance over single-model approaches, they often come at the cost of increased computational overhead, both in terms of training and inference. Recently, Model Soup~\cite{wortsman2022model} suggested interpolating the weights of a bunch of networks finetuned with different hyperparameters configurations. Also, to mitigate the overfitting of the finetuned CLIP~\cite{radford2021learning}, WiseFT~\cite{wiseft} combines the weights of fine-tuned and initial CLIP models. These approaches can attain better generalization capability without increasing the computational cost of deployed models.  

\begin{figure*}[t]
    \centering
    \includegraphics[width=0.9\linewidth]{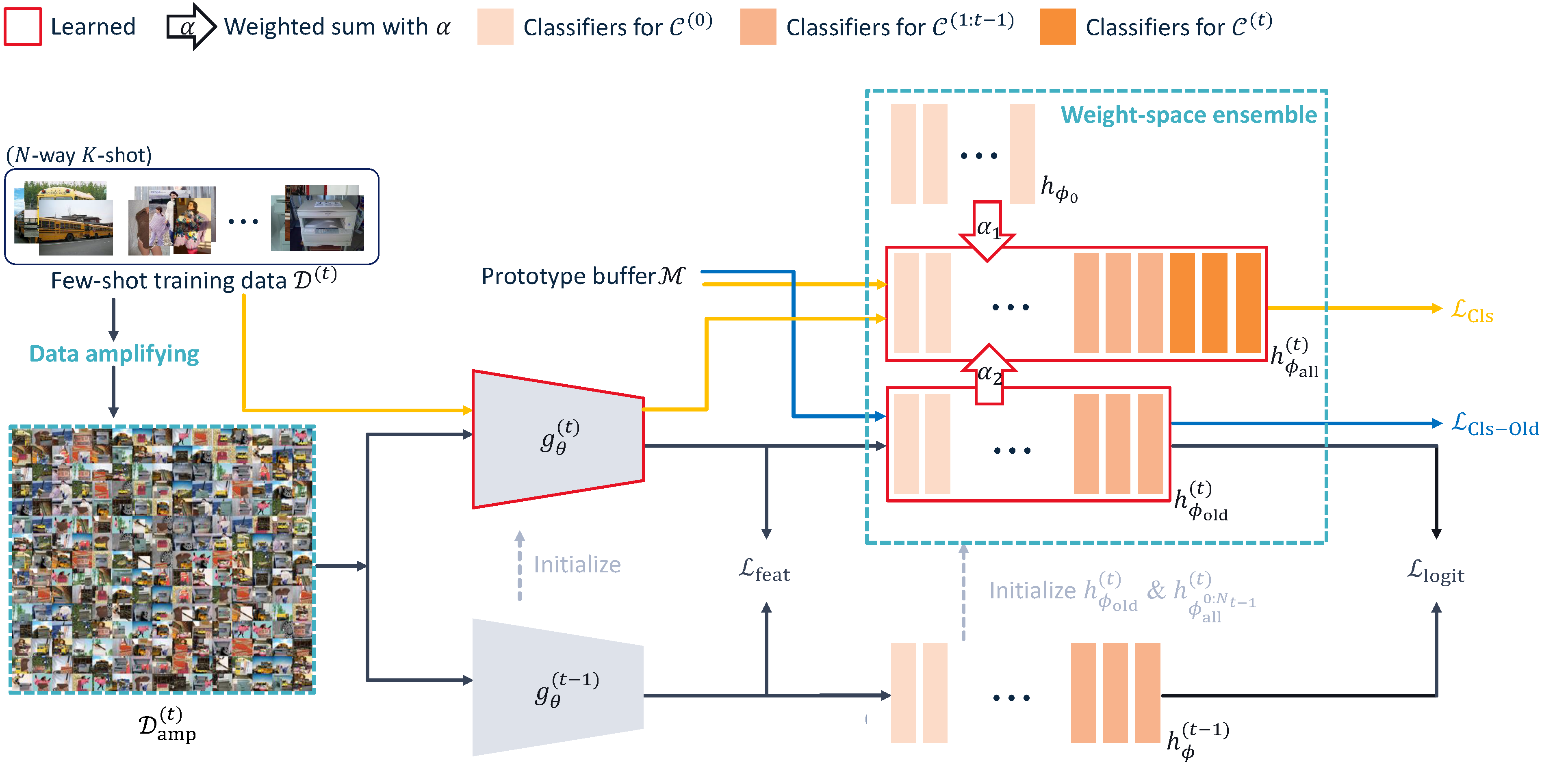}
    \caption{
    \textbf{The pipeline of the proposed method.} At the $t$th incremental session $S^{(t)}$, the model is initialized using the resulting model of session $S^{(t-1)}$. Then, a set of $N$-way $K$-shot few-shot examples $D^{(t)}$ and the prototype buffer $\mathcal{M}$ are given. The weights $\phi_0$, $\phi_{\text{old}}$, $\phi_{\text{all}}$ of classification heads $h_{\phi_{0}}$, $h_{\phi_{\text{old}}}^{(t)}$, $h_{\phi_{\text{all}}}^{(t)}$ are interpolated with learnable scalars $\alpha_1$ and $\alpha_2$. The main loss $\mathcal{L}_{\text{Cls}}$ is computed on the resulting classification head $h_{\phi}^{(t)}$'s output. Also, classification loss over the old classes, appeared until the session $S^{(t-1)}$, is computed on the output of $h_{\phi_{\text{old}}}^{(t)}$. Additionally, $D^{(t)}$ is amplified to $D^{(t)}_{\text{amp}}$, and then KD losses $\mathcal{L}_{\text{feat}}$ and $\mathcal{L}_{\text{logit}}$ are computed on the amplified dataset $D^{(t)}_{\text{amp}}$. The learned components are outlined in red. 
    }    
    \label{fig:framework}
\end{figure*}

\section{Proposed Method}
\label{sec:method}
In this section, we describe the proposed Tri-WE and ADKD for FSCIL. The pipeline is illustrated in Fig.~\ref{fig:framework}. We also provide the process of the base session training. 

\noindent\textbf{Problem setting.} FSCIL aims to incrementally scale the model's class recognition ability by learning new classes with limited training data while preserving its knowledge of previously learned, old classes. In specific, we suppose that a model $f_{\theta,\phi}$ consists of the feature extractor $g_\theta$ and the classification head (cosine classifier) $h_\phi$ where $\theta$ and $\phi$ denote learnable weights of them, respectively. Initially, the model is pre-trained on the base training set in the base session $\calS^{(0)}$. Then, the model progresses a series of $T$ incremental sessions, $\{\calS^{(1)}, \ldots, \calS^{(T)}\}$. Current session $\calS^{(t)}$ includes a class set $\calC^{(t)}$ and the corresponding training dataset $\calD^{(t)}=\{(x, y)\}$ where $x$ is a sample and $y$ is its label. 

In the base session $\calS^{(0)}$, the class set $\calC^{(0)}$ includes a substantial number of $N_0$ base classes with ample training data for each class. In contrast, in the incremental session $\calS^{(t)}$ ($t\geq1$), the class set $\calC^{(t)}$ contains $N$ classes with only a small $K$ number of training data available for every class ($N$-way $K$-shot setting). Notice that $\calC^{(i)}$ and $\calC^{(j)}$ do not overlap when $i\neq j$, $\forall i,j$. Following the prior FSCIL works~\cite{tao2020few,zhang2021few}, the model at each session $\calS^{(t)}$ only has access to $\calD^{(t)}$ and an additional prototype (the averaged feature of the $K$ examples per class) buffer $\calM$ memorizing the classes of the base and previous $(t-1)$ sessions. A single prototype is maintained for every class in $\calM$. Once the incremental learning at the session $\calS^{(t)}$ concludes, the model is evaluated on query data from all the seen $N_t$ classes thus far, that is $\calC^{(0:t)}=\calC^{(0)}\cup\calC^{(1)}\cdots\cup\calC^{(t)}$.

\subsection{Tripartite weight-space ensemble (Tri-WE)}
\label{ssec:WSE}
The base session model is generalized by numerous classes with sufficient data, and hence its weight space is conducive to adapting to new classes. Although the weight space is somewhat compromised, the intermediate previous session's model has knowledge for all the classes encountered thus far. 
To capitalize on the strengths of both models, we suggest the weight-space ensemble of three models from the sessions $\calS^{(0)}$, $\calS^{(t-1)}$, and $\calS^{(t)}$. 

However, the feature extractor, which contains many layers, is prone to overfitting or collapsing when only a few examples are available. Hence, while reducing the learning rate for the feature extractor $g^{(t)}_{\theta}$, we only ensemble the classification heads $h_{\phi_{0}}$, $h_{\phi_{\text{old}}}^{(t)}$, $h_{\phi_{\text{all}}}^{(t)}$ in the weight space. Here, $h_{\phi_{0}}$ is the classification head from the base session model. $h_{\phi_{\text{old}}}^{(t)}$ is initialized by the classification head $h_\phi^{(t-1)}$ from the session $\calS^{(t-1)}$, and then specialized to the old classes $\calC^{(0:t-1)}$. $h_{\phi_{\text{all}}}^{(t)}$ covers all the target classes $\calC^{(0:t)}$, where the old class part is initialized similarly to $h_{\phi_{\text{old}}}^{(t)}$. 
Then, as depicted in Fig.~\ref{fig:framework}, Tri-WE interpolates the classification head weight $\phi^{n}$ for $n$th class by
\begin{align}
    \label{eq:wse}
    &\phi^n=\nonumber\\
    &\begin{cases}
     \bar{\alpha}_1\phi_0^n + \bar{\alpha}_2\phi^{n}_{\text{old}} + \bar{\alpha}_3\phi_{\text{all}}^{n} \quad\;\;\;\;\text{if } n \leq N_0,&\\
     \bar{\alpha}_4\phi^{n}_{\text{old}} + \bar{\alpha}_5\phi^{n}_{\text{all}} \quad\quad\quad\;\;\,\;\;\;\quad \text{if } N_0 < n \leq N_{t-1},&\\
     \phi^{n}_{\text{all}} \quad\quad\quad\quad\quad\quad\quad\quad\quad\,\;\;\;\;\;\text{otherwise},&
    \end{cases}
\end{align}
where $1\leq n \leq N_{t}$, $\phi^n\in\mathbb{R}^{d}$, and
$d$ is the dimensionality of $\phi^n$. Then, the weight of the classification head $h_\phi^{(t)}$ to classify $N_t$ classes consists of $\{\phi^n\}_{n=1}^{N_t}$.
Notice that, we also add two learnable scalar weights $\alpha_1$ and $\alpha_2$ to automatically adjust the trade-offs between the three classification heads. Then, $\{\bar{\alpha}_1, \bar{\alpha}_2, \bar{\alpha}_3, \bar{\alpha}_4, \bar{\alpha}_5\}$ are obtained by 
\begin{align}
    \label{eq:alpha}
    \bar{\alpha}_1&= \alpha_1 / (\alpha_1 + \alpha_2 + 1),\nonumber\\
    \bar{\alpha}_2&= \alpha_2 / (\alpha_1 + \alpha_2 + 1),\nonumber\\
    \bar{\alpha}_3&= 1 / (\alpha_1 + \alpha_2 + 1),\nonumber\\
    \bar{\alpha}_4&= \alpha_2 / (\alpha_2 + 1),\nonumber\\
    \bar{\alpha}_5&= 1 / (\alpha_2 + 1).
\end{align}
When $t=1$ (i.e. $\phi_{\textrm{old}}=\phi_0$), we set $\alpha_2=0$. 

Then, the main classification loss $\mathcal{L}_\text{Cls}$ is computed on the output from the weight-space ensembled classification head $h_{\phi}^{(t)}$ by
\begin{align}
    \label{eq:loss_cls}
    \mathcal{L}_\text{Cls} = \mathbb{E}_{(x,y)\sim\calD^{(t)}}[\textrm{CE}&(h_{\phi}^{(t)}(g_{\theta}^{(t)}(x)), y] 
     \nonumber\\ & 
     +\mathbb{E}_{(\textbf{p},y)\sim\calM}[\textrm{CE}(h_{\phi}^{(t)}(\textbf{p}), y],
\end{align}
where \textrm{CE} is the cross-entropy loss function, and \textbf{p} denotes the old class prototype in $\calM$. 

Using the buffer $\mathcal{M}$, we also calculate the old class-specific loss $\mathcal{L}_\text{Cls-Old}$ for the classification head $h_{\phi_{\text{old}}}^{(t)}$ by
\begin{equation}
    \label{eq:loss_clsold}
    \mathcal{L}_\text{Cls-Old} = \mathbb{E}_{(\textbf{p},y)\sim\calM}[\textrm{CE}(h_{\phi_{\text{old}}}^{(t)}(\textbf{p}), y)].
\end{equation}
In this loss, $h_{\phi_{\text{old}}}^{(t)}$ focuses on preserving the discriminative ability regarding all the old classes on top of the current feature extractor $g_\theta^{(t)}$. Hence, we do not update $h_{\phi_{0}}$ and instead use it to retain the knowledge for the original, well-designed base class distribution from the base session model.

Unlike the naive ensemble approaches where multiple classification heads independently make predictions and combine them in the testing phase, our weight-space ensemble deploys only $h_{\phi}^{(t)}$ collaboratively utilizing the knowledge from the base and previous session models.


\subsection{Total training loss}
KD has been a widely used regularization technique in CIL to prevent the catastrophic forgetting problem~\cite{rebuffi2017icarl}. Specifically, when new classes are introduced, an additional distillation loss is combined with the main classification loss to ensure that the current model replicates the old class behavior of the previous model. Nevertheless, in FSCIL, the limited data source can lead to the extraction of knowledge biased toward those data, rather than generalized knowledge of the old classes.
To mitigate this issue, we simply generate an amplified data source $\calD_{\text{amp}}^{(t)}$ from the few-shot training data by randomly mixing pairs of data from $\calD^{(t)}$ employing simple data augmentation techniques such as~\cite{yun2019cutmix,zhang2017mixup}. The ADKD loss is then computed as
\begin{equation}
    \label{eq:loss_kd}
    \mathcal{L}_{\text{ADKD}} = \mathcal{L}_{\text{feat}} + \mathcal{L}_{\text{logit}},
\end{equation}
where $\mathcal{L}_{\text{feat}}$ and $\mathcal{L}_{\text{logit}}$ are feature-level and logit-level KD losses which are defined by
\begin{align}
    \mathcal{L}_{\text{feat}} &= \mathbb{E}_{x\sim\calD_{\text{amp}}^{(t)}} \;||g_{\theta}^{(t-1)}(x) - g_\theta^{(t)}(x)||_2 \\
    \mathcal{L}_{\text{logit}} &= \mathbb{E}_{x\sim\calD_{\text{amp}}^{(t)}} \;-f_{\theta,{\phi}}^{(t-1)}(x)\log f_{\theta,\phi^{0:N_{t-1}}}^{(t)}(x).
\end{align}
The previous session model $f_{\theta,\phi}^{(t-1)}$ including $h^{(t-1)}_{\phi}$ and $g^{(t-1)}_\theta$ is kept frozen during this process, unlike $h_{\phi_{\text{old}}}^{(t)}$.


Finally, the total loss function used to optimize the current model is
\begin{equation}
\label{eq:loss_all}
 \mathcal{L} = \mathcal{L}_{\text{Cls}} + \gamma_1\mathcal{L}_{\text{Cls-Old}} + \gamma_2\mathcal{L}_{\text{ADKD}},
\end{equation}
where $\gamma_1$ and $\gamma_2$ are hyper-parameters that balance the respective losses, $\mathcal{L}_{\text{Cls-Old}}$ and $\mathcal{L}_{\text{ADKD}}$. 
We provide the pseudo code of the overall process during an incremental session in the supplementary materials.

\subsection{Base session training}
\label{ssec:base}
In FSCIL, it is broadly acknowledged that a model with strong generalization to base classes is beneficial for learning new classes~\cite{song2023learning,yang2023neural,zhang2021few,peng2022few}. 
To ensure fairness when assessing the proposed components for incremental learning, we strive for base session performance on par with the prior SOTA methods~\cite{song2023learning,yang2023neural,zhao2023few}. For this purpose, we employ the base session training techniques of ALICE~\cite{peng2022few} (see the literature for details). On top of it, we also added a geometric classification head in parallel with the main classification head. In specific, an input is geometrically transformed into $B$ types, and we use them as the proxy labels of transformed inputs. Then, the geometric classification head is learned to classify each into one of the $B$ proxy labels. This auxiliary geometric classification helps the model encode the inherent structure of data so that the model's generalization capability on new tasks is improved. This auxiliary geometric classification strategy is not used in the incremental sessions. Through this, as aftermentioned in Sec.~\ref{sec:exp}, the base session model achieves comparable or superior performance to previous SOTA methods on three benchmark datasets. A more detailed description is in the supplementary materials.
\begin{table*}[t]
\caption{\textbf{Comparative results on miniImageNet dataset.} Session-wise Acc (\%) and the average of them (Avg) are reported. In each case, the best and second results are bold-faced and underlined, respectively. Last sess. impro. denotes the performance improvement of our method at the last session.}
\label{table:comp_mini}
\centering
\resizebox{0.9\linewidth}{!}
{
    \begin{tabular}{llccccccccccl}
    \toprule
    
    \multirow{2}{*}{Method}\; & \multirow{2}{*}{Venue}\; &\multicolumn{9}{c}{Acc in each session (\%)} & \multirow{2}{*}{Avg} \;& \multirow{2}{*}{\makecell{Last sess. \\ impro.}}\\
    \cmidrule{3-11}
    & \;& 0  \;& 1  \;&2  \;& 3  \;& 4  \;& 5  \;& 6  \;& 7  \;& 8  \;&   \;& \\
    \midrule
    \midrule
    iCaRL~\cite{rebuffi2017icarl}& CVPR 2017 \;& 61.31 \;& 46.32 \;& 42.94 \;& 37.63 \;& 30.49 \;& 24.00 \;& 20.89 \;& 18.80 \;& 17.20 \;& 33.29 \;& +43.13\\
    TOPIC~\cite{tao2020few}& CVPR 2020 \;& 61.31 \;& 50.09 \;& 45.17 \;& 41.16 \;& 37.48 \;& 35.52 \;& 32.19 \;& 29.46 \;& 24.42 \;& 39.64 \;& +35.71\\
    ERL++~\cite{dong2021few}& AAAI 2021\;& 61.70 \;& 57.58 \;& 54.66 \;& 51.72 \;& 48.66 \;& 46.27 \;& 44.67 \;& 42.81 \;& 40.79 \;& 49.87 \;& +19.34\\
    IDLVQ~\cite{chen2020incremental}& ICLR 2020\;& 64.77 \;& 59.87 \;& 55.93 \;& 52.62 \;& 49.88 \;& 47.55 \;& 44.83 \;& 43.14 \;& 41.84 \;& 51.16 \;& +18.29\\
    CEC~\cite{zhang2021few}& CVPR 2021\;& 72.00 \;& 66.83 \;& 62.97 \;& 59.43 \;& 56.70 \;& 53.73 \;& 51.19 \;& 49.24 \;& 47.63 \;& 57.74 \;& +12.50\\
    SynthFeat~\cite{cheraghian2021synthesized}& ICCV 2021\;& 61.40 \;& 59.80  \;&54.20  \;&51.69  \;&49.45 \;& 48.00  \;&45.20  \;&43.80  \;&42.10 \;& 50.63 \;& +18.03\\
    MetaFSCIL~\cite{chi2022metafscil}& CVPR 2022\;& 72.04 \;&67.94\;& 63.77 \;&60.29 \;&57.58\;& 55.16\;& 52.9\;& 50.79\;&49.19\;& 58.85 \;& +10.94\\
    FACT~\cite{zhou2022forward}& CVPR 2022 \;& 72.56 \;&69.63 \;&66.38 \;&62.77\;& 60.60 \;&57.33 \;&54.34 \;&52.16\;& 50.49 \;& 60.70 \;& +9.64\\
    Replay~\cite{liu2022few}& ECCV 2022\;& 71.84 \;&67.12 \;&63.21 \;&59.77 \;&57.01 \;&53.95 \;&51.55 \;&49.52 \;&48.21 \;&58.02 \;& +11.92\\
    ALICE~\cite{peng2022few}& ECCV 2022 \;& 80.60 \;&70.60 \;&67.40 \;&64.50 \;&62.50 \;&60.00 \;&57.80 \;&56.80 \;&55.70 \;&63.99 \;& +4.43\\
    S3C~\cite{kalla2022s3c}& ECCV 2022 \;& 76.55 \;&71.74 \;&67.66 \;&64.52 \;&61.51 \;&58.09 \;&55.36 \;&53.04 \;&51.34 \;&62.20 \;& +8.79\\
    WaRP~\cite{kim2022warping}& ICLR 2023 \;& 72.99 \;& 68.10 \;& 64.31 \;& 61.30 \;& 58.64 \;& 56.08 \;& 53.40 \;& 51.72 \;&50.65 \;&59.69 \;& +9.48\\
    SoftNet~\cite{kang2022soft}& ICLR 2023 \;&76.63 \;&70.13 \;&65.92 \;&62.52 \;&59.49 \;&56.56 \;&53.71 \;&51.72 \;&50.48 \;&60.79 \;& +9.65\\
    NC-FSCIL~\cite{yang2023neural}& ICLR 2023\;& \underline{84.02} \;& \underline{76.80} \;&72.00 \;&67.83 \;&66.35 \;& \underline{64.04} \;& \underline{61.46} \;& \underline{59.54} \;& \underline{58.31} \;& \underline{67.82} \;& +1.82\\
    GKEAL~\cite{zhuang2023gkeal}& CVPR 2023\;& 73.59 \;& 68.90 \;& 65.33 \;& 62.29 \;& 59.39 \;& 56.70 \;& 54.20 \;& 52.59 \;& 51.31\;& 60.48 \;& +10.40\\
    BiDistill~\cite{zhao2023few}& CVPR 2023\;& 74.65 \;&70.43 \;&66.29 \;&62.77 \;&60.75 \;&57.24 \;&54.79 \;&53.65 \;&52.22 \;&61.42 \;& +7.91\\
    SAVC~\cite{song2023learning}& CVPR 2023\;& 81.12 \;&76.14 \;&\underline{72.43} \;&\underline{68.92} \;& \underline{66.48} \;&62.95 \;&59.92 \;&58.39 \;&57.11 \;&67.05 \;& +3.02\\
    OrCo~\cite{ahmed2024orco}& CVPR 2024\;& 83.30 \;&75.32 \;&71.52 \;&68.16 \;& 65.62 \;&63.11 \;&60.20 \;&58.82 \;&58.08 \;&67.12 \;& +2.05\\
    CLOSER~\cite{oh2024closer} & ECCV 2024\;& 76.02 \;&71.61 \;&67.99 \;&64.69 \;&61.70 \;&58.94 \;&56.23 \;&54.52 \;&53.33 \;&62.78 \;& +6.80\\
    \midrule 
    Ours \;& \;&\textbf{84.13} \;&\textbf{81.41} \;&\textbf{76.65} \;&\textbf{73.59} \;&\textbf{70.10} \;&\textbf{65.13} \;&\textbf{63.42} \;&\textbf{61.02} \;&\textbf{60.13} \;&\textbf{70.62} \;& \\ 

    \bottomrule
    \end{tabular}
}
\end{table*}

\begin{figure*}[t]
  \centering
    
   \includegraphics[width=0.72\linewidth]{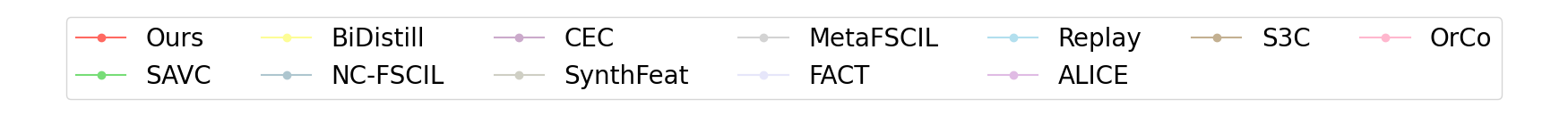} \\
   \begin{subfigure}{0.48\linewidth}\includegraphics[width=\linewidth]{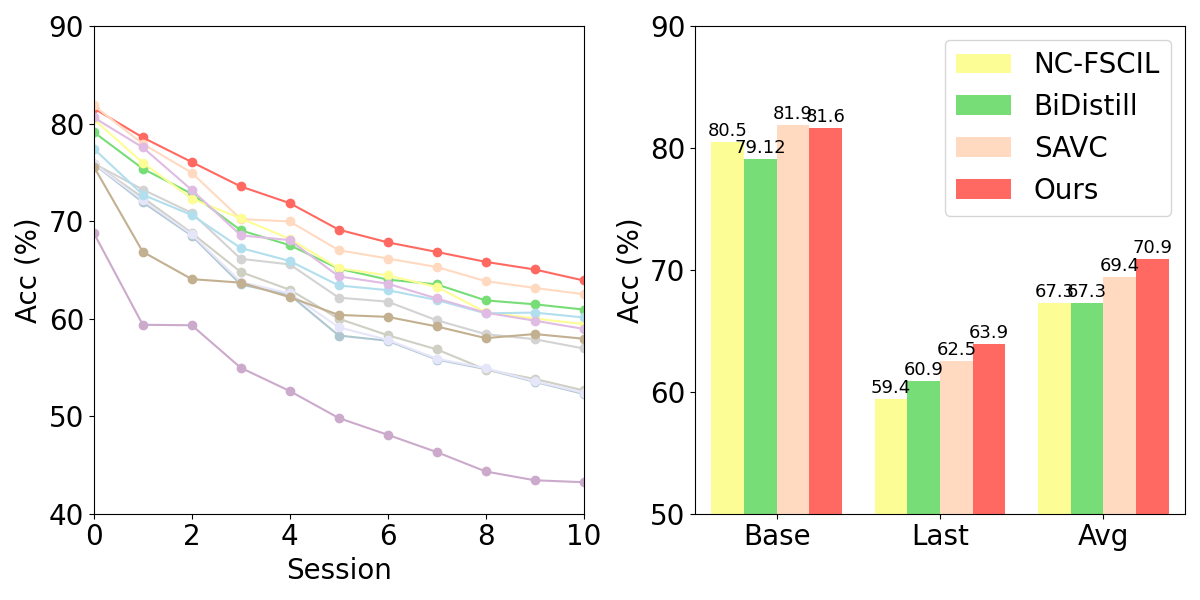}
    \caption{CUB200}
    \end{subfigure} \begin{subfigure}{0.48\linewidth}\includegraphics[width=\linewidth]{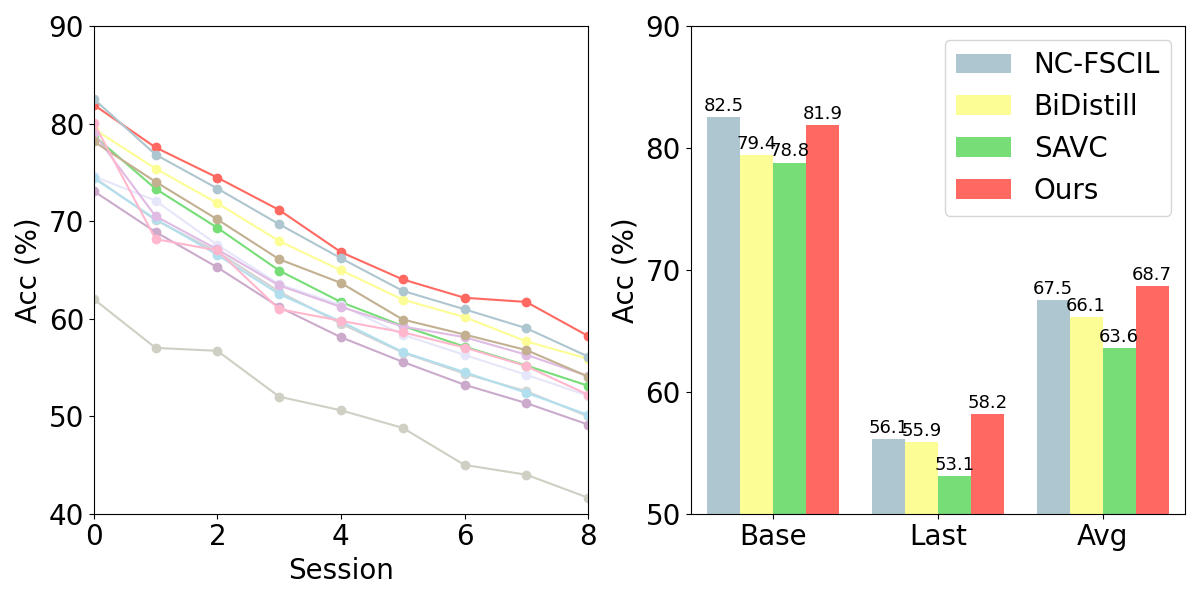}
    \caption{CIFAR100}
    \end{subfigure}

   
  \caption{\textbf{Comparative results on CUB200 and CIFAR100 datasets.} For each dataset, the curve of the session-wise Acc (\%) is first provided. And also, the base, last, average Acc are reported on our method comparing with three recent SOTA methods, NC-FSCIL~\cite{yang2023neural}, SAVC~\cite{song2023learning}, YourSelf~\cite{tang2024rethinking} (Best viewed in color).} 
  \label{fig:plot_comp}
\end{figure*}

\section{Experimental Results}
\label{sec:exp}

\noindent\textbf{Datasets.} Consistent with the setting of~\cite{tao2020few}, we conduct our experiments on three benchmark datasets: miniImageNet~\cite{matchingnet}, CIFAR100~\cite{krizhevsky2009learning} and CUB200~\cite{wah2011caltech}. miniImageNet is a subset of ImageNet~\cite{deng2009imagenet} that includes 60,000 images from 100 classes, each with resolution 84$\times$84. CIFAR100 comprises 60,000 tiny images of size 32$\times$32 across 100 categories. CUB200, which is focused on a fine-grained classification, contains 200 bird species with similar appearance where the image size is 224$\times$224. We follow the splits from~\cite{tao2020few}. For miniImageNet and CIFAR100, 60 categories are selected as base classes while the remaining are split into 8 incremental sessions with only 5 training examples per novel class (i.e., 5-way 5-shot). As for the CUB200 dataset, 100 categories are designated as the base training set, while the rest are divided into 10-way 5-shot tasks across 10 sessions.

\noindent\textbf{Evaluation protocol.} In line with the FSCIL literature, we employ the top-1 accuracy (Acc) for all the observed classes at the end of each incremental learning session. We also provide the average of Acc over all the sessions. 

\noindent\textbf{Implementation details.}
We implement our method using the PyTorch library. In keeping with the previous works~\cite{kang2022soft,peng2022few,zhao2023few,song2023learning}, ResNet18~\cite{he2016deep} serves as our feature extractor. The entire model is trained via the SGD optimizer. For the base session, the learning rates are initialized by 0.01, 0.001, and 0.01 for miniImageNet, CUB200, and CIFAR100 respectively, and decayed by 0.1 at 60 and 70 epochs. In each incremental session, the learning rate for the weight-space ensembled $\phi_{\text{all}}$ is 0.1, while it is 0.001 for the rest. Hence, the feature extractor is minimally updated in incremental learning. For the ADKD loss, the given $NK$ training examples are amplified to $16NK$ for each class. Also, the learnable scalars $\alpha_1$ and $\alpha_2$ are initially set to 1.0. Loss weights $\lambda_1$ and $\lambda_2$ are empirically determined to be 1.2 and 10.0, respectively. Further details are in Sec. S-5 of the supplementary materials.

\begin{table*}[t]
\caption{\textbf{Ablation analysis for Tri-WE.} Varying the combination of classification heads ensembled in weight space (No WE, Dual-WE, Tri-WE), session-wise and average Acc (\%) are provided on miniImageNet. Also, ``Naive'' means that none of WE and ADKD is used.}
\label{table:we_ablation}
\centering
\renewcommand{\arraystretch}{1.0}
\resizebox{0.95\linewidth}{!}
{
    \begin{tabular}{lcccccccccccccccccc}
    \toprule
    & \multicolumn{7}{c}{Classification head weight} && \multicolumn{9}{c}{Acc in each session (\%)} & \multirow{2}{*}{Avg}\\
    \cmidrule{2-8} \cmidrule{10-18}
    &\quad\quad& $\phi_0$  &\quad\quad\quad& $\phi_{\text{old}}$  &\quad\quad\quad& $\phi_{\text{all}}$  &\quad& & 0 & 1 & 2 & 3 & 4 & 5 & 6 & 7 & 8 &  \\
    
    \midrule
    \midrule
    Naive && && && &&  & 84.13\;&80.55\;&71.26\;&60.93\;&50.05\;&44.08\;&36.42\;&19.78\;&16.73\;&51.55 \\
    \midrule
    No WE&& && && $\checkmark$ &&  & 
    84.13\;&78.95\;&74.09\;&70.06\;&66.56\;&63.07\;&60.32\;&58.08\;&56.01\;&67.92\\
    \midrule
    \multirow{2}{*}{Dual-WE}&& && $\checkmark$ && $\checkmark$ &&  & 
    84.13\;&80.25\;&75.47\;&71.58\;&67.62\;&64.52\;&61.11\;&60.17\;&58.93\;&69.31\\
    &&$\checkmark$ && && $\checkmark$ &&  & 
    84.13\;&79.54\;&74.99\;&70.13\;&66.60\;&63.43\;&60.85\;&59.89\;&57.31\;&68.54\\
    \midrule
    \textbf{Tri-WE} &&$\checkmark$ && $\checkmark$ && $\checkmark$ && &84.13\;&81.41\;&76.65\;&73.59\;&70.1\;&65.13\;&63.42\;&61.02\;&60.13\;&70.62\\
    
    \bottomrule
    \end{tabular}
}
\end{table*}

\begin{table*}[t]
\caption{\textbf{Impact on increasing the layers ensembled in weight space.} Applying the proposed Tri-WE to the convolutional blocks of the feature extractor (ResNet18), session-wise and average Acc (\%) are provided on miniImageNet.}
\label{table:head_ablation}
\centering
\renewcommand{\arraystretch}{1.0}
\resizebox{0.8\linewidth}{!}
{
    \begin{tabular}{lcccccccccccc}
    \toprule
    \multirow{2}{*}{Weight-space ensembled} && \multicolumn{9}{c}{Acc in each session (\%)} && \multirow{2}{*}{Avg}\\
    
    \cmidrule{3-11}
    && 0 & 1 & 2 & 3 & 4 & 5 & 6 & 7 & 8 && \\
    
    \midrule
    Blocks 1-4 && 
    84.13\;&75.25\;&70.25\;&67.54\;&64.85\;&61.34\;&59.74\;&56.38\;&53.45\;&&65.88\\
    Blocks 1-4 + clf. head && 84.13\;&76.08\;&72.13\;&68.42\;&65.83\;&62.60\;&60.19\;&57.75\;&55.29\;&&66.94\\
    Blocks 2-4 + clf. head  &&84.13\;&77.84\;&73.42\;&69.22\;&65.85\;&62.75\;&60.56\;&57.34\;&55.33\;&&67.38\\
    Blocks 3-4 + clf. head  &&84.13\;&78.59\;&74.43\;&69.80\;&66.25\;&63.13\;&60.46\;&58.04\;&55.65\;&&67.83\\
    Block 4 + clf. head &&84.13\;&78.80\;&74.21\;&69.64\;&66.01\;&63.32\;&61.50\;&58.17\;&55.88\;&&67.96\\
    \midrule
    \textbf{Clf. head only }&&84.13\;&81.41\;&76.65\;&73.59\;&70.1\;&65.13\;&63.42\;&61.02\;&60.13\;&&70.62\\
    
    \bottomrule
    \end{tabular}
}
\end{table*}

\begin{figure}[t]
    \centering
    \begin{subfigure}{0.5\linewidth}\includegraphics[width=\linewidth]{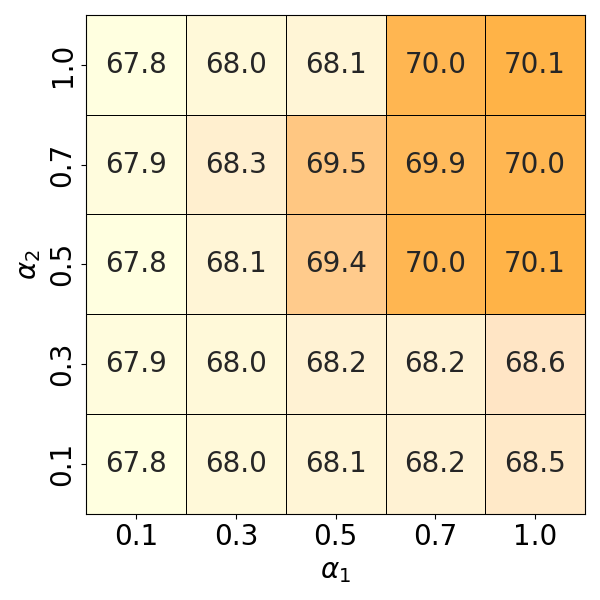}
    \caption{Fixed ($\alpha_1$, $\alpha_2$)}
    \end{subfigure}
    \begin{subfigure}{0.42\linewidth}\includegraphics[width=\linewidth]{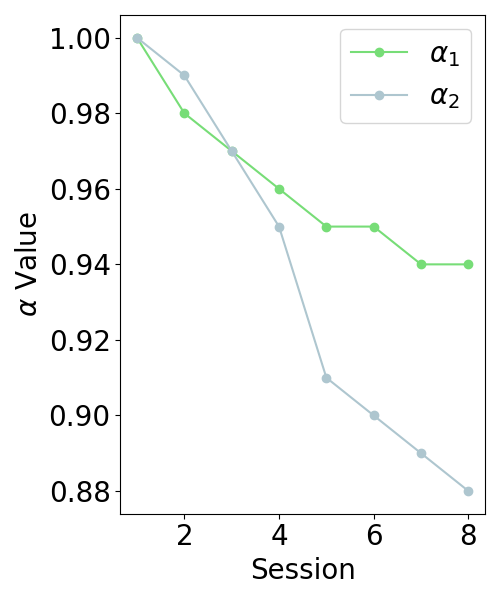}
    \caption{Learned $\alpha_1$, $\alpha_2$}
    \end{subfigure}
    \caption{
    \textbf{Analysis on learnable scalars $\alpha_1$ and $\alpha_2$  on miniImageNet. } (a) Comparison with multiple combinations of fixed (not-learnable) $\alpha_1$ and $\alpha_2$ in terms of averaged Acc over all the sessions, (b) session-wise learned $\alpha_1$ and $\alpha_2$ in ours. 
    }    
    \label{fig:detail}
\end{figure}

\subsection{Comparative assessment}
On the three benchmark datasets, we first compare the proposed method with the existing approaches. As indicated in Table~\ref{table:comp_mini} on the miniImageNet dataset, our method outperforms the compared in all the sessions, achieving a clear SOTA in terms of the average Acc by a margin of 1.82\% at least. 
Also, BiDistill and SAVC require the base session model and a considerable number of classifiers per class, respectively, even in the testing phase. NC-FSCIL and OrCo benefit from knowing the total number of incremental classes at the outset of the base session.  Whereas, we attain clear SOTA with a smaller-scale deployed model compared to BiDistill and SAVC, and without the information advantage of NC-FSCIL and OrCo. 

Moreover, Fig.~\ref{fig:plot_comp} shows the Acc curve according to sessions on the CUB200 and CIFAR100 datasets. As in the miniImageNet dataset, our method consistently surpasses the compared methods by large gaps in the incremental sessions on the CUB 200 dataset. In specific, despite lower base session Acc than SAVC, our method outperforms it in the last session. Also, while NC-FSCIL merely finetuned the last module of the feature encoder in incremental sessions, OrCo attempted to also optimize the pre-set classifiers. Nevertheless, the additional update of only the classifiers is less effective. On the CIFAR100 dataset, we can see the similar tendency. NC-FSCIL’s setting is somewhat favorable, as it allows the pre-definition of classifiers for the incremental classes with knowledge of all class numbers in advance at the base session. This advantage is particularly beneficial on the CIFAR100 dataset, where some classes are semantically similar (e.g., possum and mouse), and the low resolution (32$\times$32) of the images could otherwise complicate classification. The detailed results are available in Sec. S-4 of the supplementary materials. These findings demonstrate our method's superior ability to adapt to new classes with limited training data while maintaining seamless integration with the old classes.

\subsection{Ablation analysis}
We extensively analyze the components of the proposed method on the miniImageNet or CUB200 datasets.

\noindent\textbf{Tripartite strategy.}
We conduct the ablation experiments on the Tri-WE in Table~\ref{table:we_ablation}. Before this, initially, we contrast ``Naive'' with the ``No WE'' configuration, the latter incorporating the ADKD loss. The naive approach, lacking both WE and ADKD loss, exhibits a significant performance decline due to simultaneous updates of the feature extractor and classification head. However, No WE achieves respectable results thanks to the ADKD loss (more discussed in the following sections). Next, as demonstrated in Table~\ref{table:we_ablation}, the absence of the weight-space ensemble for the classification head leads to increased catastrophic forgetting. And, from ``Dual-WE,'' we can see that the model from the intermediate previous session is more important than the base session model, as the reduced learning rate for the feature extractor mitigates drastic changes, helping to preserve the integrity of the feature space.
Accordingly, the knowledge for incrementally-appeared classes is more helpful. Lastly, the complete Tri-WE clearly outperforms No WE by 2.70\% in average Acc, effectively retaining the essential knowledge of the base and previous session models.

\noindent\textbf{WE in classification head.}
We also evaluate the effect of weight-space ensemble varying the number of interpolated layers in Table \ref{table:head_ablation}. In specific, we apply the proposed Tri-WE on the convolutional blocks of the feature extractors. Here, we use different learnable scalars depending on the convolutional blocks, rather than sharing them across the convolutional blocks and classification head. We observed a tendency for the decline in performance as the number of weight-space-ensembled layers increases. This is because that deciding how to merge a lot of layers is difficult with only a few examples. Also, only merging feature extractor (1st row of Table \ref{table:head_ablation}) shows the lowest performance. It emphasizes that the head's weight-space is crucial to prevent catastrophic forgetting.

\begin{table}[tb]
\caption{\textbf{Analysis on data amplification schemes.} The last and average Acc (\%) are reported on each scheme on miniImageNet.}
\label{table:adkd}
\centering
\renewcommand{\arraystretch}{1.0}
\resizebox{\linewidth}{!}
{
    \begin{tabular}{cccccc}
    \toprule
    & \multirow{2}{*}{w/o AD} &\multicolumn{4}{c}{AD scheme} \\
    \cmidrule{3-6}
    & & CutOut~\cite{devries2017improved} & RandAug~\cite{cubuk2020randaugment} & MixUp~\cite{zhang2017mixup} & CutMix~\cite{yun2019cutmix} \\
    \midrule
    Last Acc& 58.08 & 58.12 & 58.26 & 59.14 & 60.13\\
    Avg Acc& 69.05 & 69.13 & 69.09 & 70.03 & 70.62\\
    \bottomrule
    \end{tabular}
}
\end{table}

\noindent\textbf{Learnable scalars $\alpha_1$\&$\alpha_2$.}
Fig.~\ref{fig:detail}(a) shows the averaged Acc on diverse combinations of $(\alpha_1, \alpha_2)$ where both are fixed after initialization in all the sessions. Compared to all the fixed cases, the learnable scalars show better result (70.62\% in the averaged Acc). Also, as Fig.~\ref{fig:detail}(b), we show the learned $\alpha_1$ and $\alpha_2$ in each session in our method where we initialize both as 1.0 in every session. We can see that the learned $\alpha_1$ and $\alpha_2$ are different depending on sessions. Accordingly, our learnable scalar approach is useful for balancing the trade-off among the three classification heads. We also provide session-wise results on the fixed cases in the supplementary materials.

\noindent\textbf{Effect of data amplification.} 
While Tri-WE already attains SOTA (i.e. w/o AD in Table~\ref{table:adkd}), incorporating our ADKD loss provides further increase (69.05\% \textit{vs} 70.62\% in average Acc). Also, to see the influence of the choice of data amplification scheme in the ADKD loss, we compare multiple data augmentation methods for our data amplification. To facilitate the use of ADKD in follow-up works, we consider four widely-used data augmentation methods, CutOut~\cite{devries2017improved}, RandAug~\cite{cubuk2020randaugment}, CutMix~\cite{yun2019cutmix}, MixUp~\cite{zhang2017mixup}. In the first two, the images are not inter-mixed. Whereas, two images are mixed for data augmentation. In Table~\ref{table:adkd}, we can see that CutOut and RandAug are not effective compared to ``w/o AD'' where no data amplification is considered. With only a few examples, those methods cannot yield a diversity of data. Next, in comparison with MixUp, CutMix shows a better result. MixUp mixes two images smoothly, and then the original semantics are somewhat preserved. This is less useful in KD. Namely, CutMix encourages the model to focus more on general image recognition ability rather than high-level semantics, which is more helpful in preventing the catastrophic forgetting problem in FSCIL. Hence, we employ CutMix as the default option. In addition, the impact of the degree of data amplification is addressed in Sec. S-1 of the supplementary materials.

\noindent\textbf{Two loss terms of $\mathcal{L}_\text{ADKD}$.}
As shown in the first two rows of Table~\ref{table:loss}, we ablate $\mathcal{L}_\text{feat}$ and $\mathcal{L}_\text{logit}$, respectively. From this study, we can see that $\mathcal{L}_\text{feat}$ is slightly more important than $\mathcal{L}_\text{logit}$ since Tri-WE already helps to prevent catastrophic forgetting in the side of the classification head. Then, using both is most effective. 

\noindent\textbf{$\mathcal{L}_\text{Cls-Old}$.} As we update the feature extractor as well as the classification head in the incremental learning, we also adapt $\phi_{\text{old}}$ to the updated feature extractor using the loss $\mathcal{L}_\text{Cls-Old}$. To verify this, we conduct ablating the feature extractor update and $\mathcal{L}_\text{Cls-Old}$. As shown in Table~\ref{table:loss_clsold}, the performance is degraded without this loss term when the feature extractor is updated. Also, when the feature extractor is frozen, updating $\phi_{\text{old}}$ is less meaningful. Finally, updating the feature extractor with $\phi_{\text{old}}$ is most beneficial.

\begin{table}[t]
\caption{\textbf{Analysis on loss terms of $\mathcal{L}_{\text{ADKD}}$.} The last and average Acc (\%) are reported ablating the loss terms on miniImageNet.}
\label{table:loss}
\centering
\renewcommand{\arraystretch}{1.0}
\resizebox{0.5\linewidth}{!}
{
    \begin{tabular}{cc|cc}
    \toprule
    $\mathcal{L}_{\text{feat}}$ & $\mathcal{L}_{\text{logit}}$ & Last Acc & Avg Acc\\
    \midrule
    \checkmark& & 59.42 & 70.01 \\
    & \checkmark & 58.77 & 68.41 \\
    \checkmark& \checkmark& 60.13& 70.62 \\
    \bottomrule
    \end{tabular}
}
\end{table}

\begin{table}[t]
\caption{\textbf{Analysis on $\mathcal{L}_{\text{Cls-Old}}$ and updating the feature extractor $g_\theta$.} The last and average Acc (\%) are reported, on miniImageNet.}
\label{table:loss_clsold}
\centering
\renewcommand{\arraystretch}{1.0}
\resizebox{0.55\linewidth}{!}
{
    \begin{tabular}{cc|cc}
    \toprule
    $\mathcal{L}_{\text{Cls-Old}}$ & Update $g_\theta$ & Last Acc & Avg Acc\\
    \midrule
    & \checkmark & 59.27& 70.15  \\
    \checkmark&\checkmark& 60.13& 70.62 \\
    \midrule
    & & 58.86& 69.41  \\
    \checkmark& & 58.88& 69.40  \\
    \bottomrule
    \end{tabular}
}
\vspace{0.3cm}
\end{table}

\begin{figure}[t]
    \centering
    \includegraphics[width=\linewidth]{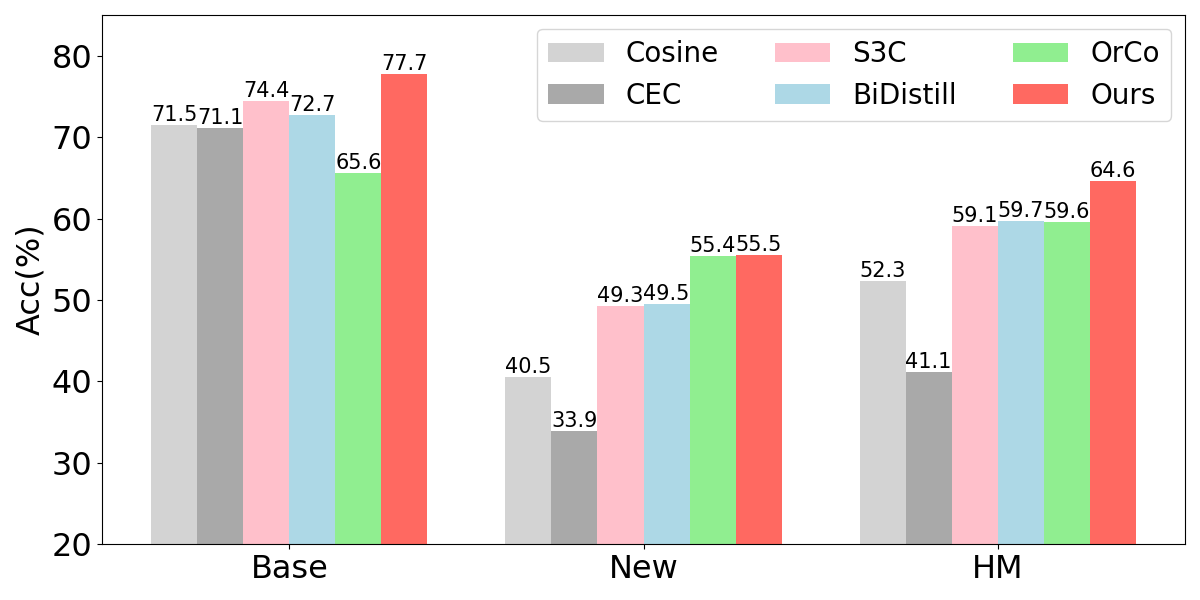}
    \caption{
    \textbf{Analysis on the new class adaptability on CUB200.} The average of accuracy scores are provided for base and novel classes, denoted by Base and New respectively. And, harmonic means are also compared.}    
    \vspace{-0.5cm}
    \label{fig:hm}
\end{figure}

\noindent{\textbf{New class adaptability.}} 
To see the efficacy of our method in terms of the trade-off between base and newly introduced classes, we provide the individual accuracy of the base and novel classes, and the harmonic mean (HM). Since most previous works have addressed the joint accuracy of the base and novel classes, we can compare our method with only a few recent methods~\cite{zhou2022forward,zhang2021few,zhao2023few,kalla2022s3c,matchingnet} on the CUB200 dataset, following~\cite{zhou2022forward,kalla2022s3c}. Fig.~\ref{fig:hm} shows that our approach is more adept at integrating new classes without compromising the performance on the base classes. 

\section{Conclusions}
\label{con}
We explored FSCIL which facilitates the continuous acquisition of new concepts from a few examples. To this end, we suggested a novel FSCIL method including Tri-WE and ADKD regularization loss. The proposed method can effectively prevent overfitting to the few data and catastrophic forgetting of old concepts in the model. Tri-WE strategically interpolates the weights of the classification heads from the base, immediate previous, and current models, which collectively preserves the knowledge from the earlier models without overwhelming updates. Enriching the few examples for KD, the ADKD loss facilitates the transfer of essential knowledge from the previous model, rather than the knowledge overfitted to the few examples. We conducted extensive analyses on the proposed components and achieved state-of-the-art performance on the miniImageNet, CUB200, and CIFAR100 datasets.

{
    \small
    \bibliographystyle{ieeenat_fullname}
    \bibliography{FSCIL_BIB}

\begin{thebibliography}{47}
\providecommand{\natexlab}[1]{#1}
\providecommand{\url}[1]{\texttt{#1}}
\expandafter\ifx\csname urlstyle\endcsname\relax
  \providecommand{\doi}[1]{doi: #1}\else
  \providecommand{\doi}{doi: \begingroup \urlstyle{rm}\Url}\fi

\bibitem[Ahmed et~al.(2024)Ahmed, Kukleva, and Schiele]{ahmed2024orco}
Noor Ahmed, Anna Kukleva, and Bernt Schiele.
\newblock Orco: Towards better generalization via orthogonality and contrast for few-shot class-incremental learning.
\newblock In \emph{IEEE Conf. Comput. Vis. Pattern Recog.}, 2024.

\bibitem[Belouadah and Popescu(2019)]{belouadah2019il2m}
Eden Belouadah and Adrian Popescu.
\newblock Il2m: Class incremental learning with dual memory.
\newblock In \emph{IEEE Conf. Comput. Vis. Pattern Recog.}, 2019.

\bibitem[Chaudhry et~al.(2018)Chaudhry, Ranzato, Rohrbach, and Elhoseiny]{chaudhry2018efficient}
Arslan Chaudhry, Marc'Aurelio Ranzato, Marcus Rohrbach, and Mohamed Elhoseiny.
\newblock Efficient lifelong learning with a-gem.
\newblock \emph{arXiv preprint arXiv:1812.00420}, 2018.

\bibitem[Chen and Lee(2020)]{chen2020incremental}
Kuilin Chen and Chi-Guhn Lee.
\newblock Incremental few-shot learning via vector quantization in deep embedded space.
\newblock In \emph{Int. Conf. Learn. Represent.}, 2020.

\bibitem[Cheraghian et~al.(2021)Cheraghian, Rahman, Ramasinghe, Fang, Simon, Petersson, and Harandi]{cheraghian2021synthesized}
Ali Cheraghian, Shafin Rahman, Sameera Ramasinghe, Pengfei Fang, Christian Simon, Lars Petersson, and Mehrtash Harandi.
\newblock Synthesized feature based few-shot class-incremental learning on a mixture of subspaces.
\newblock In \emph{Int. Conf. Comput. Vis.}, 2021.

\bibitem[Chi et~al.(2022)Chi, Gu, Liu, Wang, Yu, and Tang]{chi2022metafscil}
Zhixiang Chi, Li Gu, Huan Liu, Yang Wang, Yuanhao Yu, and Jin Tang.
\newblock Metafscil: A meta-learning approach for few-shot class incremental learning.
\newblock In \emph{IEEE Conf. Comput. Vis. Pattern Recog.}, 2022.

\bibitem[Cubuk et~al.(2020)Cubuk, Zoph, Shlens, and Le]{cubuk2020randaugment}
Ekin~D Cubuk, Barret Zoph, Jonathon Shlens, and Quoc~V Le.
\newblock Randaugment: Practical automated data augmentation with a reduced search space.
\newblock In \emph{IEEE Conf. Comput. Vis. Pattern Recog. Worksh.}, 2020.

\bibitem[Deng et~al.(2009)Deng, Dong, Socher, Li, Li, and Fei-Fei]{deng2009imagenet}
Jia Deng, Wei Dong, Richard Socher, Li-Jia Li, Kai Li, and Li Fei-Fei.
\newblock Imagenet: A large-scale hierarchical image database.
\newblock In \emph{IEEE Conf. Comput. Vis. Pattern Recog.}, 2009.

\bibitem[DeVries and Taylor(2017)]{devries2017improved}
Terrance DeVries and Graham~W Taylor.
\newblock Improved regularization of convolutional neural networks with cutout.
\newblock \emph{arXiv preprint arXiv:1708.04552}, 2017.

\bibitem[Dietterich(2000)]{dietterich2000ensemble}
Thomas~G Dietterich.
\newblock Ensemble methods in machine learning.
\newblock In \emph{International Workshop on Multiple Classifier Systems}, 2000.

\bibitem[Dong et~al.(2021)Dong, Hong, Tao, Chang, Wei, and Gong]{dong2021few}
Songlin Dong, Xiaopeng Hong, Xiaoyu Tao, Xinyuan Chang, Xing Wei, and Yihong Gong.
\newblock Few-shot class-incremental learning via relation knowledge distillation.
\newblock In \emph{AAAI}, 2021.

\bibitem[Douillard et~al.(2020)Douillard, Cord, Ollion, Robert, and Valle]{douillard2020podnet}
Arthur Douillard, Matthieu Cord, Charles Ollion, Thomas Robert, and Eduardo Valle.
\newblock Podnet: Pooled outputs distillation for small-tasks incremental learning.
\newblock In \emph{Eur. Conf. Comput. Vis.}, 2020.

\bibitem[Hansen and Salamon(1990)]{hansen1990neural}
Lars~Kai Hansen and Peter Salamon.
\newblock Neural network ensembles.
\newblock \emph{IEEE Trans. Pattern Anal. Mach. Intell.}, 12\penalty0 (10):\penalty0 993--1001, 1990.

\bibitem[He et~al.(2016)He, Zhang, Ren, and Sun]{he2016deep}
Kaiming He, Xiangyu Zhang, Shaoqing Ren, and Jian Sun.
\newblock Deep residual learning for image recognition.
\newblock In \emph{IEEE Conf. Comput. Vis. Pattern Recog.}, 2016.

\bibitem[Hou et~al.(2019)Hou, Pan, Loy, Wang, and Lin]{hou2019learning}
Saihui Hou, Xinyu Pan, Chen~Change Loy, Zilei Wang, and Dahua Lin.
\newblock Learning a unified classifier incrementally via rebalancing.
\newblock In \emph{IEEE Conf. Comput. Vis. Pattern Recog.}, 2019.

\bibitem[Huang et~al.(2017)Huang, Li, Pleiss, Liu, Hopcroft, and Weinberger]{huang2017iclr}
Gao Huang, Yixuan Li, Geoff Pleiss, Zhuang Liu, John~E Hopcroft, and Kilian~Q Weinberger.
\newblock Snapshot ensembles{: T}rain 1, get m for free.
\newblock In \emph{Int. Conf. Learn. Represent.} 2017.

\bibitem[Kalla and Biswas(2022)]{kalla2022s3c}
Jayateja Kalla and Soma Biswas.
\newblock S3c: Self-supervised stochastic classifiers for few-shot class-incremental learning.
\newblock In \emph{Eur. Conf. Comput. Vis.}, 2022.

\bibitem[Kang et~al.(2023)Kang, Yoon, Madjid, Hwang, and Yoo]{kang2022soft}
Haeyong Kang, Jaehong Yoon, Sultan Rizky~Hikmawan Madjid, Sung~Ju Hwang, and Chang~D Yoo.
\newblock On the soft-subnetwork for few-shot class incremental learning.
\newblock In \emph{Int. Conf. Learn. Represent.}, 2023.

\bibitem[Kim et~al.(2023)Kim, Han, Seo, and Moon]{kim2022warping}
Do-Yeon Kim, Dong-Jun Han, Jun Seo, and Jaekyun Moon.
\newblock Warping the space: Weight space rotation for class-incremental few-shot learning.
\newblock In \emph{Int. Conf. Learn. Represent.}, 2023.

\bibitem[Krizhevsky et~al.(2009)Krizhevsky, Hinton, et~al.]{krizhevsky2009learning}
Alex Krizhevsky, Geoffrey Hinton, et~al.
\newblock Learning multiple layers of features from tiny images.
\newblock 2009.

\bibitem[Krizhevsky et~al.(2012)Krizhevsky, Sutskever, and Hinton]{krizhevsky2012imagenet}
Alex Krizhevsky, Ilya Sutskever, and Geoffrey~E Hinton.
\newblock Imagenet classification with deep convolutional neural networks.
\newblock In \emph{Adv. Neural Inform. Process. Syst.} 2012.

\bibitem[Li and Hoiem(2017)]{li2017learning}
Zhizhong Li and Derek Hoiem.
\newblock Learning without forgetting.
\newblock \emph{IEEE Trans. Pattern Anal. Mach. Intell.}, 40\penalty0 (12):\penalty0 2935--2947, 2017.

\bibitem[Liu et~al.(2022)Liu, Gu, Chi, Wang, Yu, Chen, and Tang]{liu2022few}
Huan Liu, Li Gu, Zhixiang Chi, Yang Wang, Yuanhao Yu, Jun Chen, and Jin Tang.
\newblock Few-shot class-incremental learning via entropy-regularized data-free replay.
\newblock In \emph{Eur. Conf. Comput. Vis.}, 2022.

\bibitem[Oh et~al.(2024)Oh, Baik, and Lee]{oh2024closer}
Junghun Oh, Sungyong Baik, and Kyoung~Mu Lee.
\newblock Closer: Towards better representation learning for few-shot class-incremental learning.
\newblock In \emph{Eur. Conf. Comput. Vis.}, 2024.

\bibitem[Peng et~al.(2022)Peng, Zhao, Wang, Li, and Lovell]{peng2022few}
Can Peng, Kun Zhao, Tianren Wang, Meng Li, and Brian~C Lovell.
\newblock Few-shot class-incremental learning from an open-set perspective.
\newblock In \emph{Eur. Conf. Comput. Vis.}, 2022.

\bibitem[Radford et~al.(2021)Radford, Kim, Hallacy, Ramesh, Goh, Agarwal, Sastry, Askell, Mishkin, Clark, et~al.]{radford2021learning}
Alec Radford, Jong~Wook Kim, Chris Hallacy, Aditya Ramesh, Gabriel Goh, Sandhini Agarwal, Girish Sastry, Amanda Askell, Pamela Mishkin, Jack Clark, et~al.
\newblock Learning transferable visual models from natural language supervision.
\newblock In \emph{Int. Conf. Machine Learning}, 2021.

\bibitem[Rebuffi et~al.(2017)Rebuffi, Kolesnikov, Sperl, and Lampert]{rebuffi2017icarl}
Sylvestre-Alvise Rebuffi, Alexander Kolesnikov, Georg Sperl, and Christoph~H Lampert.
\newblock icarl: Incremental classifier and representation learning.
\newblock In \emph{IEEE Conf. Comput. Vis. Pattern Recog.}, 2017.

\bibitem[Rolnick et~al.(2019)Rolnick, Ahuja, Schwarz, Lillicrap, and Wayne]{rolnick2019experience}
David Rolnick, Arun Ahuja, Jonathan Schwarz, Timothy Lillicrap, and Gregory Wayne.
\newblock Experience replay for continual learning.
\newblock 2019.

\bibitem[Song et~al.(2023)Song, Zhao, Shi, Peng, Yuan, and Tian]{song2023learning}
Zeyin Song, Yifan Zhao, Yujun Shi, Peixi Peng, Li Yuan, and Yonghong Tian.
\newblock Learning with fantasy: Semantic-aware virtual contrastive constraint for few-shot class-incremental learning.
\newblock In \emph{IEEE Conf. Comput. Vis. Pattern Recog.}, 2023.

\bibitem[Szegedy et~al.(2015)Szegedy, Liu, Jia, Sermanet, Reed, Anguelov, Erhan, Vanhoucke, and Rabinovich]{szegedy2015going}
Christian Szegedy, Wei Liu, Yangqing Jia, Pierre Sermanet, Scott Reed, Dragomir Anguelov, Dumitru Erhan, Vincent Vanhoucke, and Andrew Rabinovich.
\newblock Going deeper with convolutions.
\newblock In \emph{IEEE Conf. Comput. Vis. Pattern Recog.} 2015.

\bibitem[Tang et~al.(2024)Tang, Peng, Meng, and Zheng]{tang2024rethinking}
Yu-Ming Tang, Yi-Xing Peng, Jingke Meng, and Wei-Shi Zheng.
\newblock Rethinking few-shot class-incremental learning: Learning from yourself.
\newblock In \emph{Eur. Conf. Comput. Vis.}, 2024.

\bibitem[Tao et~al.(2020)Tao, Hong, Chang, Dong, Wei, and Gong]{tao2020few}
Xiaoyu Tao, Xiaopeng Hong, Xinyuan Chang, Songlin Dong, Xing Wei, and Yihong Gong.
\newblock Few-shot class-incremental learning.
\newblock In \emph{IEEE Conf. Comput. Vis. Pattern Recog.}, 2020.

\bibitem[Vinyals et~al.(2016)Vinyals, Blundell, Lillicrap, Kavukcuoglu, and Wierstra]{matchingnet}
Oriol Vinyals, Charles Blundell, Tim Lillicrap, Koray Kavukcuoglu, and Daan Wierstra.
\newblock Matching networks for one shot learning.
\newblock In \emph{NeurIPS}, 2016.

\bibitem[Wah et~al.(2011)Wah, Branson, Welinder, Perona, and Belongie]{wah2011caltech}
Catherine Wah, Steve Branson, Peter Welinder, Pietro Perona, and Serge Belongie.
\newblock The caltech-ucsd birds-200-2011 dataset.
\newblock 2011.

\bibitem[Wang et~al.(2022)Wang, Zhang, Lee, Zhang, Sun, Ren, Su, Perot, Dy, and Pfister]{wang2022learning}
Zifeng Wang, Zizhao Zhang, Chen-Yu Lee, Han Zhang, Ruoxi Sun, Xiaoqi Ren, Guolong Su, Vincent Perot, Jennifer Dy, and Tomas Pfister.
\newblock Learning to prompt for continual learning.
\newblock In \emph{IEEE Conf. Comput. Vis. Pattern Recog.}, 2022.

\bibitem[Wortsman et~al.(2022{\natexlab{a}})Wortsman, Ilharco, Gadre, Roelofs, Gontijo-Lopes, Morcos, Namkoong, Farhadi, Carmon, Kornblith, et~al.]{wortsman2022model}
Mitchell Wortsman, Gabriel Ilharco, Samir~Ya Gadre, Rebecca Roelofs, Raphael Gontijo-Lopes, Ari~S Morcos, Hongseok Namkoong, Ali Farhadi, Yair Carmon, Simon Kornblith, et~al.
\newblock Model soups: averaging weights of multiple fine-tuned models improves accuracy without increasing inference time.
\newblock In \emph{Int. Conf. Machine Learning}, 2022{\natexlab{a}}.

\bibitem[Wortsman et~al.(2022{\natexlab{b}})Wortsman, Ilharco, Kim, Li, Kornblith, Roelofs, Lopes, Hajishirzi, Farhadi, Namkoong, and Schmidt]{wiseft}
Mitchell Wortsman, Gabriel Ilharco, Jong~Wook Kim, Mike Li, Simon Kornblith, Rebecca Roelofs, Raphael~Gontijo Lopes, Hannaneh Hajishirzi, Ali Farhadi, Hongseok Namkoong, and Ludwig Schmidt.
\newblock Robust fine-tuning of zero-shot models.
\newblock In \emph{IEEE Conf. Comput. Vis. Pattern Recog.}, 2022{\natexlab{b}}.

\bibitem[Wu et~al.(2019)Wu, Chen, Wang, Ye, Liu, Guo, and Fu]{wu2019large}
Yue Wu, Yinpeng Chen, Lijuan Wang, Yuancheng Ye, Zicheng Liu, Yandong Guo, and Yun Fu.
\newblock Large scale incremental learning.
\newblock In \emph{IEEE Conf. Comput. Vis. Pattern Recog.}, 2019.

\bibitem[Yan et~al.(2021)Yan, Xie, and He]{yan2021dynamically}
Shipeng Yan, Jiangwei Xie, and Xuming He.
\newblock Der: Dynamically expandable representation for class incremental learning.
\newblock In \emph{IEEE Conf. Comput. Vis. Pattern Recog.}, 2021.

\bibitem[Yang et~al.(2023)Yang, Yuan, Li, Lin, Torr, and Tao]{yang2023neural}
Yibo Yang, Haobo Yuan, Xiangtai Li, Zhouchen Lin, Philip Torr, and Dacheng Tao.
\newblock Neural collapse inspired feature-classifier alignment for few-shot class incremental learning.
\newblock In \emph{Int. Conf. Learn. Represent.}, 2023.

\bibitem[Yun et~al.(2019)Yun, Han, Oh, Chun, Choe, and Yoo]{yun2019cutmix}
Sangdoo Yun, Dongyoon Han, Seong~Joon Oh, Sanghyuk Chun, Junsuk Choe, and Youngjoon Yoo.
\newblock Cutmix: Regularization strategy to train strong classifiers with localizable features.
\newblock In \emph{Int. Conf. Comput. Vis.}, 2019.

\bibitem[Zhang et~al.(2021)Zhang, Song, Lin, Zheng, Pan, and Xu]{zhang2021few}
Chi Zhang, Nan Song, Guosheng Lin, Yun Zheng, Pan Pan, and Yinghui Xu.
\newblock Few-shot incremental learning with continually evolved classifiers.
\newblock In \emph{IEEE Conf. Comput. Vis. Pattern Recog.}, 2021.

\bibitem[Zhang et~al.(2017)Zhang, Cisse, Dauphin, and Lopez-Paz]{zhang2017mixup}
Hongyi Zhang, Moustapha Cisse, Yann~N Dauphin, and David Lopez-Paz.
\newblock mixup: Beyond empirical risk minimization.
\newblock \emph{arXiv preprint arXiv:1710.09412}, 2017.

\bibitem[Zhao et~al.(2020)Zhao, Xiao, Gan, Zhang, and Xia]{zhao2020maintaining}
Bowen Zhao, Xi Xiao, Guojun Gan, Bin Zhang, and Shu-Tao Xia.
\newblock Maintaining discrimination and fairness in class incremental learning.
\newblock In \emph{IEEE Conf. Comput. Vis. Pattern Recog.}, 2020.

\bibitem[Zhao et~al.(2023)Zhao, Lu, Xu, Cheng, Guo, Niu, and Fang]{zhao2023few}
Linglan Zhao, Jing Lu, Yunlu Xu, Zhanzhan Cheng, Dashan Guo, Yi Niu, and Xiangzhong Fang.
\newblock Few-shot class-incremental learning via class-aware bilateral distillation.
\newblock In \emph{IEEE Conf. Comput. Vis. Pattern Recog.}, 2023.

\bibitem[Zhou et~al.(2022)Zhou, Wang, Ye, Ma, Pu, and Zhan]{zhou2022forward}
Da-Wei Zhou, Fu-Yun Wang, Han-Jia Ye, Liang Ma, Shiliang Pu, and De-Chuan Zhan.
\newblock Forward compatible few-shot class-incremental learning.
\newblock In \emph{IEEE Conf. Comput. Vis. Pattern Recog.}, 2022.

\bibitem[Zhuang et~al.(2023)Zhuang, Weng, He, Lin, and Zeng]{zhuang2023gkeal}
Huiping Zhuang, Zhenyu Weng, Run He, Zhiping Lin, and Ziqian Zeng.
\newblock Gkeal: Gaussian kernel embedded analytic learning for few-shot class incremental task.
\newblock In \emph{IEEE Conf. Comput. Vis. Pattern Recog.}, 2023.

\end{thebibliography}
}


\end{document}